\documentclass[letterpaper]{article} % DO NOT CHANGE THIS
\usepackage{aaai2026}  % DO NOT CHANGE THIS
\usepackage{times}  % DO NOT CHANGE THIS
\usepackage{helvet}  % DO NOT CHANGE THIS
\usepackage{courier}  % DO NOT CHANGE THIS
\usepackage[hyphens]{url}  % DO NOT CHANGE THIS
\usepackage{graphicx} % DO NOT CHANGE THIS
\usepackage{natbib}  % DO NOT CHANGE THIS AND DO NOT ADD ANY OPTIONS TO IT
\usepackage{caption} % DO NOT CHANGE THIS AND DO NOT ADD ANY OPTIONS TO IT
\usepackage{algorithm}
\usepackage{algorithmic}

\usepackage{newfloat}
\usepackage{listings}
\DeclareCaptionStyle{ruled}{labelfont=normalfont,labelsep=colon,strut=off} % DO NOT CHANGE THIS
\floatstyle{ruled}
\newfloat{listing}{tb}{lst}{}
\floatname{listing}{Listing}
\title{AlignCVC: Aligning Cross-View Consistency for Single-Image-to-3D Generation}

\author{
  Xinyue Liang\thanks{These authors contributed equally.}, 
  Zhiyuan Ma\footnotemark[1], 
  Lingchen Sun, 
  Yanjun Guo, 
  Lei Zhang\thanks{Corresponding author.}
}

\affiliations{
  Department of Computing, The Hong Kong Polytechnic University\\
  Hong Kong, China\\
  \texttt{\{xinyue.liang, zm2354.ma, ling-chen.sun, yanjunn.guo\}@connect.polyu.hk, cslzhang@polyu.edu.hk}
}

\usepackage{bibentry}
\usepackage{algorithm}
\usepackage{algorithmic}
\usepackage{booktabs}
\usepackage{makecell}
\usepackage{graphicx}
\usepackage{multicol}
\usepackage{xcolor}
\usepackage{caption}
\definecolor{darkgreen}{RGB}{0,100,0} % 定义深绿色
\usepackage{pifont}

\usepackage{minted}

\usepackage{newfloat}
\usepackage{listings}

\usepackage{amsmath}
\usepackage{amssymb} % 支持 \triangleq 和 \mathbb

\begin{document}

\maketitle

\vspace{-7mm}

\begin{abstract}

Single-image-to-3D models typically follow a sequential generation and reconstruction workflow. However, intermediate multi-view images synthesized by pre-trained generation models often lack cross-view consistency (CVC), significantly degrading 3D reconstruction performance. While recent methods attempt to refine CVC by feeding reconstruction results back into the multi-view generator, these approaches struggle with noisy and unstable reconstruction outputs that limit effective CVC improvement.
We introduce AlignCVC, a novel framework that fundamentally re-frames single-image-to-3D generation through distribution alignment rather than relying on strict regression losses. Our key insight is to align both generated and reconstructed multi-view distributions toward the ground-truth multi-view distribution, establishing a principled foundation for improved CVC. Observing that generated images exhibit weak CVC while reconstructed images display strong CVC due to explicit rendering, we propose a soft-hard alignment strategy with distinct objectives for generation and reconstruction models. This approach not only enhances generation quality but also dramatically accelerates inference to as few as 4 steps.
As a plug-and-play paradigm, our method, namely AlignCVC, seamlessly integrates various combinations of multiview generation models with 3D reconstruction models. Extensive experiments demonstrate the effectiveness and efficiency of AlignCVC for single-image-to-3D generation.

\end{abstract}  

\begin{links}
    \link{Code}{https://github.com/Liangsanzhu/AlignCVC.git}
\end{links}

\section{Introduction}

\begin{figure*}[t]
\centering
\vspace{-5mm}

  \includegraphics[width=0.85\linewidth]{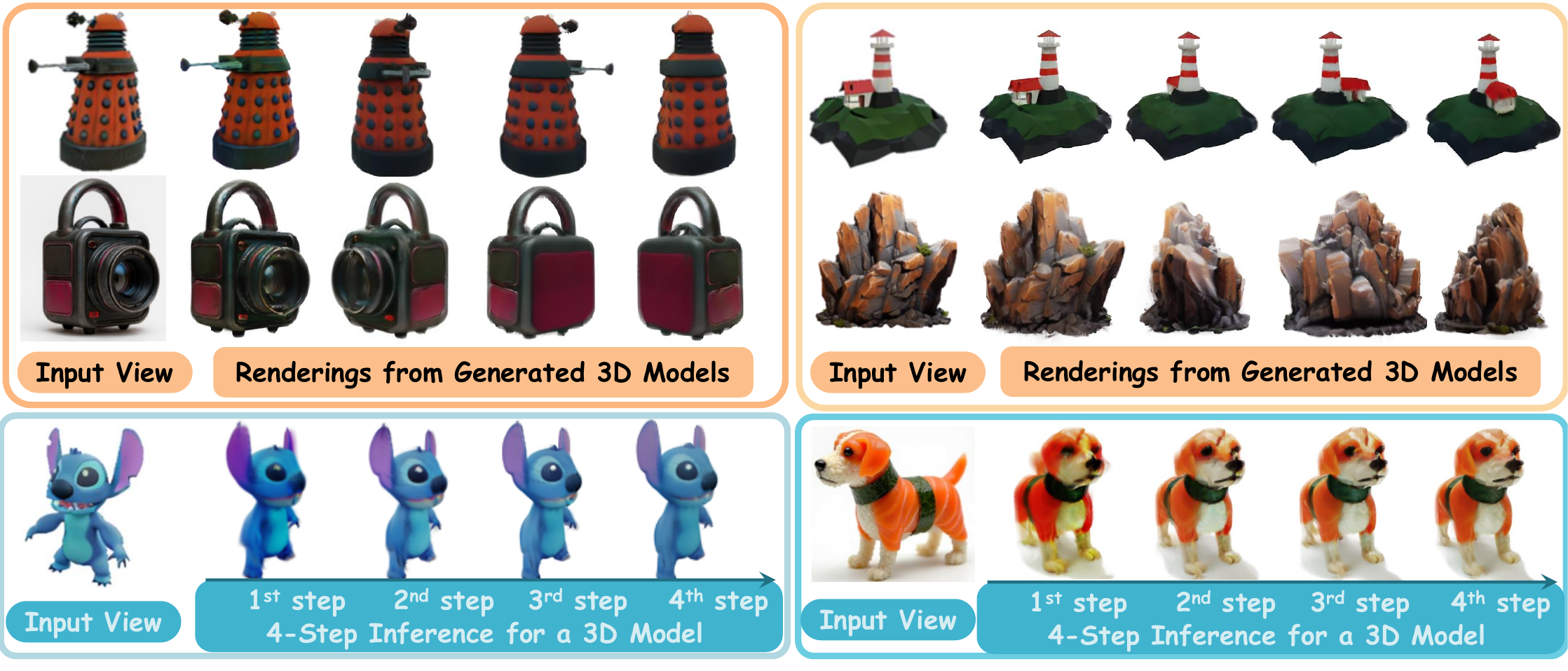}
\vspace{-4mm}

\caption{Our AlignCVC method jointly post-trains multi-view generation and reconstruction models with \textbf{distribution alignment} for 3D-aware sampling, enabling high-fidelity image-to-3D generation with only 4 diffusion steps for efficient inference.}    
\label{fig:teaser}     
\vspace{-6mm}
\end{figure*}

With advancements in 3D datasets and large-scale pre-trained 2D diffusion models, single-image-to-3D generation~\cite{liu2023zero,liu2024one,liu2024one_plus,liu2023syncdreamer,long2024wonder3d,tang2025lgm} has made significant progress in recent years. One approach leverages 3D priors from 3D datasets for 3D generation~\cite{zhang2024clay,xiang2024structured,zhao2025hunyuan3d}. However, these methods struggle with generalization and flexibility due to the limited size and diversity of existing 3D datasets~\cite{deitke2023objaverse,deitke2024objaverse,collins2022abo,fu20213d}, which are smaller and less varied than their 2D counterparts, restricting applicability across domains.
Another approach utilizes pre-trained 2D diffusion models, such as Stable Diffusion~\cite{rombach2022high} and Stable Video Diffusion~\cite{blattmann2023stable}, to generate 3D content. Early works rely on Score Distillation~\cite{poole2022dreamfusion,wang2024prolificdreamer,ma2025scaledreamer,yu2023text} to iteratively optimize a single 3D model for a given image, which is time-consuming. Recent works employ multi-view generation (MVG) models~\cite{liu2023zero,long2024wonder3d,wang2023imagedream,huang2024epidiff} to synthesize sparse-view images, followed by feed-forward 3D reconstruction~\cite{liu2023zero,chen2025lara,hong2023lrm,tang2025lgm,xu2024grm} for greater efficiency.
% \begin{figure}[!t]

%     \centering
%     \includegraphics[width=1\linewidth]{figs/flow.pdf}
%     \vspace{-5mm}
%     \caption{\textbf{Comparison among different multi-view generation (MVG) based image-to-3D generation paradigms}. While existing methods typically require $T \geq 25$ steps with low CVC, we effectively enhance the CVC with only 4-step sampling by aligning the MVG and the reconstructed outputs to the target distribution.}    
%     \vspace{-3mm}
%     \label{fig:flow}                             
% \end{figure}

However, pre-trained MVG models often struggle with cross-view consistency (CVC) when synthesizing sparse-view images, posing significant challenges for 3D reconstruction and undermining the reliability of the final 3D output. Recent works~\cite{melas20243d,chen20243d,zuo2024videomv,wen2024ouroboros3d,xue2024human,tang2024cycle3d,xie2024carve3d,xue2024gen} attempt to enhance CVC through 3D-aware sampling, which feeds rendered 3D reconstructions back into the generation stage. However, these methods lack effective joint optimization between generation and reconstruction models. While noisy intermediate multi-view images result in low-quality reconstructions, the integration of generation and reconstruction outputs remains fragile, as they are only optimized against fixed GT images, ignoring the inherent diversity of generation distribution under a certain input~\cite{chen20243d,zuo2024videomv,wen2024ouroboros3d,xue2024human,tang2024cycle3d,xue2024gen}. Additionally, 3D-aware sampling methods are slow~\cite{wen2024ouroboros3d,zuo2024videomv,xue2024gen,xue2024human,tang2024cycle3d}, often requiring over 25 recursive steps, limiting their practicality and efficiency.

%inspired by the recent advances in score distillation techniques~\cite{poole2022dreamfusion,wang2024prolificdreamer,yu2023text,ma2025scaledreamer},%
To address these, we propose relaxing the rigid constraints of input-conditioned generation and reconstruction. Unlike previous methods, where each input corresponds to fixed target multi-views, our approach aligns the generation process with a distribution derived from images rendered from 3D assets. By aligning both generation and reconstruction within a shared distribution space, we minimize discrepancies between them and enable more stable 3D-aware sampling.
Specifically, we introduce a soft-hard alignment strategy: soft alignment for generation models to enhance implicit CVC, and hard alignment for reconstruction models to enforce explicit CVC. This distribution-aligned approach produces cleaner intermediate images and more stable outputs with higher CVC, outperforming 3D-aware sampling methods relying solely on regression losses with GT images.

%Notably, certain generation and reconstruction models are inherently equipped to integrate geometric priors. Upon this, we further incorporate geometry alongside appearance information into the reconstruction and sampling processes, fostering a more profound synergy that elevates CVC.

As illustrated in Fig.~\ref{fig:teaser}, our proposed paradigm, namely \textbf{AlignCVC}, serves as a plug-and-play solution for \textbf{Align}ing the \textbf{CVC} for various generation-reconstruction based image-to-3D pipelines. 
Extensive experiments on representative diffusion models such as Wonder3D~\cite{long2024wonder3d} and SV3D~\cite{voleti2024sv3d}, together with reconstruction models such as LGM~\cite{tang2025lgm}, GeoLRM~\cite{zhang2024geolrm}, and LaRa~\cite{chen2025lara}, validate its robustness and adaptability.
Our contributions are as follows:

\begin{itemize}
    \item We propose a plug-and-play paradigm, AlignCVC, for single-image-to-3D generation that aligns MVG and reconstruction results with the distribution rendered from high-quality 3D assets, enabling more effective 3D-aware sampling to enhance the robustness of CVC.
    
    \item We propose a soft-hard distribution alignment strategy, assigning a soft objective to MVG models and a hard objective to reconstruction models.

%     \item We further incorporate geometric information into reconstruction and sampling, thereby achieving additional improvements in CVC.

    \item Extensive experiments show that our plug-and-play paradigm boosts CVC and enhances image-to-3D generation quality across various MVG and reconstruction models. Additionally, we cut multi-view diffusion steps to just 4, achieving much faster inference speed than other 3D-aware sampling methods.

\end{itemize}

\section{Related Work}

\textbf{Score Distillation for Generation}.  
Score Distillation Sampling (SDS)\cite{poole2022dreamfusion} transfers 2D generation knowledge into 3D generation by aligning 3D renderings with distributions from pre-trained diffusion models. As a base for recent advancements in 3D generation~\cite{lin2023magic3d,shi2023mvdream,wang2023score,yu2023text, wang2024prolificdreamer,liu2023zero,ma2025progressive,ma2025scaledreamer}, SDS leverages pre-trained diffusion models effectively. However, its dependence on per-scene optimization limits inference speed and generalizability, making it unsuitable for real-time 3D generation and driving the development of more efficient feed-forward methods.

\textbf{Multi-view Generation (MVG).} MVG generates multiple images simultaneously to enhance 3D generation speed and consistency. Recent models~\cite{shi2023mvdream,liu2023syncdreamer,liu2023zero,huang2024epidiff,long2024wonder3d,blattmann2023stable} fine-tune pre-trained diffusion models on large 3D datasets~\cite{deitke2023objaverse} for multi-view image generation. Stable Video Diffusion~\cite{blattmann2023stable} supports multi-frame outputs, with which SV3D~\cite{voleti2024sv3d} generating orbit-view videos from a single image. Integrating MVG improves both efficiency and quality in 3D pipelines.  
Feed-forward reconstruction models~\cite{chen2025lara,tang2025lgm,zhang2024geolrm} address this by accelerating sparse-view reconstruction, forming a two-stage paradigm (see Fig.~\ref{fig:flow} (a)). However, the lack of explicit cross-view consistency (CVC) constraints in MVG-generated views often degrades reconstruction performance.

\textbf{3D-Aware Sampling}. 3D-aware-sampling-based methods~\cite{chen20243d,wen2024ouroboros3d,xue2024gen,zuo2024videomv,tang2024cycle3d} enhance MVG's CVC by integrating feedback from reconstructed 3D renderings and explicit CVC constraints into the generation process (see Fig.~\ref{fig:flow} (b)). While improving CVC, these methods face challenges: noisy intermediate images (see Fig.~\ref{fig:noise_image}) degrade final 3D quality, weak joint optimization between generation and reconstruction limits performance, and multi-step recursion (25–50 iterations) hinders time efficiency.

\section{Method}
\label{sec:method}

\begin{figure*}[!t]

    \centering
    \includegraphics[width=0.85\linewidth]{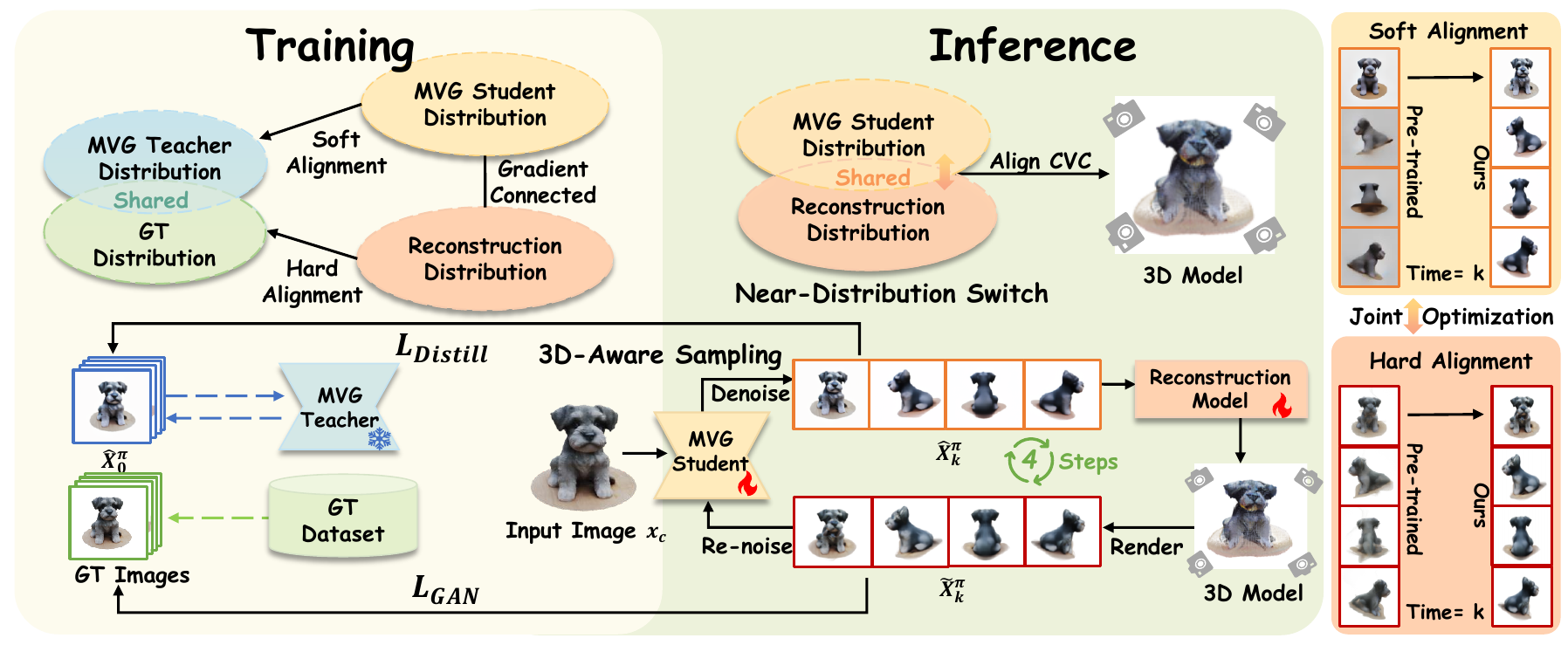}
    \vspace{-3mm}
    \caption{\textbf{The framework of AlignCVC.} During training, the multi-view generation (MVG) student model generates multi-view images $\hat{\boldsymbol{X}}^{\pi}_{k}$  from input image $\boldsymbol{x}_c$ at camera poses $\pi$. A pre-trained MVG teacher model aligns $\hat{\boldsymbol{X}}^{\pi}_{k}$ with the GT distribution via a soft-alignment method. We then obtain the 3D model from the reconstruction model and adversarially supervise its renderings $\Tilde{\boldsymbol{X}}^{\pi}_{k}$ to the GT distribution in a hard-aligned manner. In the inference phase, we reconstruct an intermediate 3D model with the generated multi-view images $\hat{\boldsymbol{X}}^{\pi}_{k}$ at each timestep, where the renderings $\Tilde{\boldsymbol{X}}^{\pi}_{k}$ are then re-noised for the next denoising timestep with 3D-aware sampling. This recursive sampling, repeated for 4 steps, produces the final 3D model.
    }     
    \label{fig:pipeline_image}              
         
    \vspace{-5mm}    
\end{figure*}

In this paper, we propose \textbf{AlignCVC}, a novel framework that enhances CVC by shifting strict regression losses with distribution alignment against GT, enabling more efficient and higher-quality 3D model generation through 3D-aware sampling.
 As shown in Fig.~\ref{fig:pipeline_image}, AlignCVC bridges generation and reconstruction through a 3D-aware sampling framework, progressively refining the final 3D model by incorporating 3D priors. Specifically, the input image is first processed by an MVG model using Score Distillation for soft alignment, generating intermediate multi-views aligned with the GT distribution. These multi-views are then refined by a reconstruction model with a hard alignment strategy, where adversarial learning ensures the renderings of the reconstructed 3D model remain consistent with the GT distribution. By integrating the implicit CVC of MVG with the explicit CVC of reconstruction, AlignCVC efficiently delivers a high-quality and consistent final 3D model.

\subsection{Preliminaries and Motivation}
\label{subsec:preliminary}

\textbf{MVG Model}. In this paper, we focus on diffusion-based MVG models. Specifically, the MVG takes a single view $\boldsymbol{x}_c$ as the input condition and generates $N$ views $\hat{\boldsymbol{X}}^{\pi} = \{ \boldsymbol{x}^{\pi_1}, \boldsymbol{x}^{\pi_2}, \dots, \boldsymbol{x}^{\pi_N} \}$ corresponding to camera poses $\pi=\{\pi_1, \dots, \pi_N\}$ by initializing from Gaussian distribution $\mathcal{N}(0,\boldsymbol{I})$ and progressively denoising in $K$ steps $t_{K} > t_{K-1} > \dots > t_1$. The denoising results at the last timestep $t_1$ are taken as the final generation results $\hat{\boldsymbol{X}}^{\pi}_{0}$.

\begin{figure}[t]
\centering
    \vspace{-4mm}    

  \includegraphics[width=0.9\linewidth]{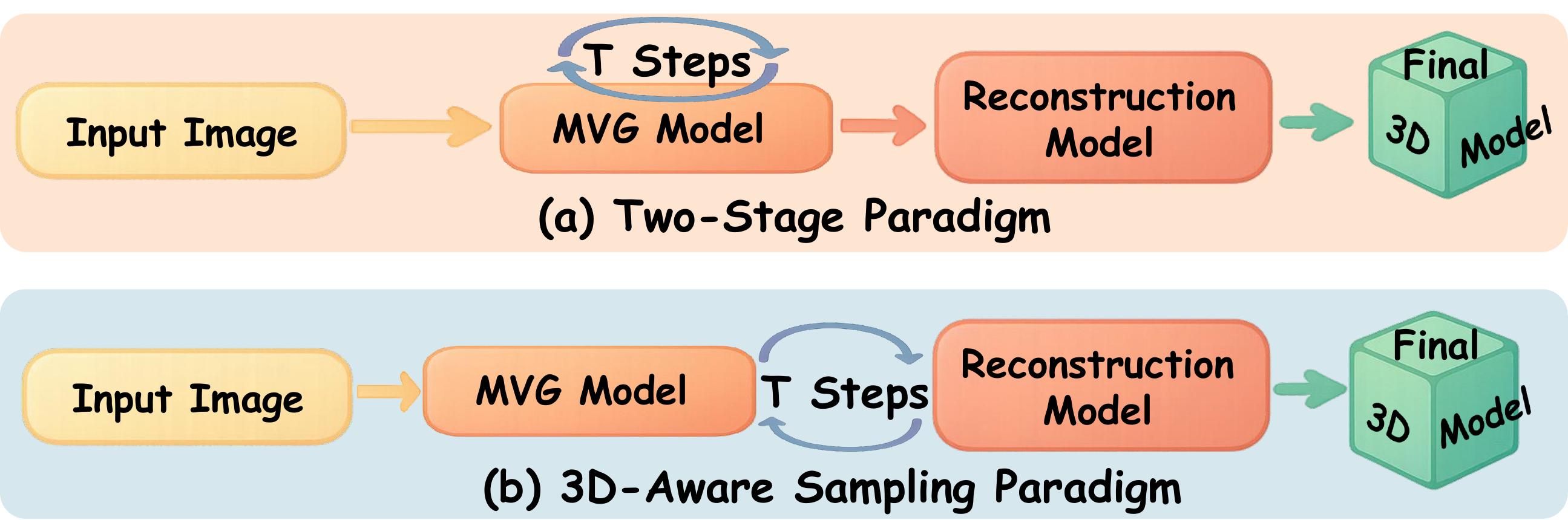}
  
\vspace{-3mm}

\caption{Two typical MVG-based Image-to-3D approaches.}%(a) Two-stage generation-reconstruction paradigm. (b) 3D-aware sampling paradigm.}    
\vspace{-7mm}
\label{fig:flow}                             
\end{figure}

\textbf{Reconstruction Model}. Given the generated multi-views $\hat{\boldsymbol{X}}^{\pi}_{0}$, the reconstruction model produces a 3D output, differentiably rendered as $\Tilde{\boldsymbol{X}}^{\pi}_{0}$ at camera poses $\pi$. This two-stage generation-reconstruction framework is shown in Fig.~\ref{fig:flow} (a). However, since most reconstruction models are trained only on GT multi-views with perfect CVC and rely heavily on CVC, such as aggregating multi-view features in 3D space, they often fail on noisy and inconsistent views from MVG, producing a low-quality 3D model misaligned with $\boldsymbol{x}_c$.
\begin{figure}[h]
    \centering
        \vspace{-5mm}    

    \includegraphics[width=0.9\linewidth]{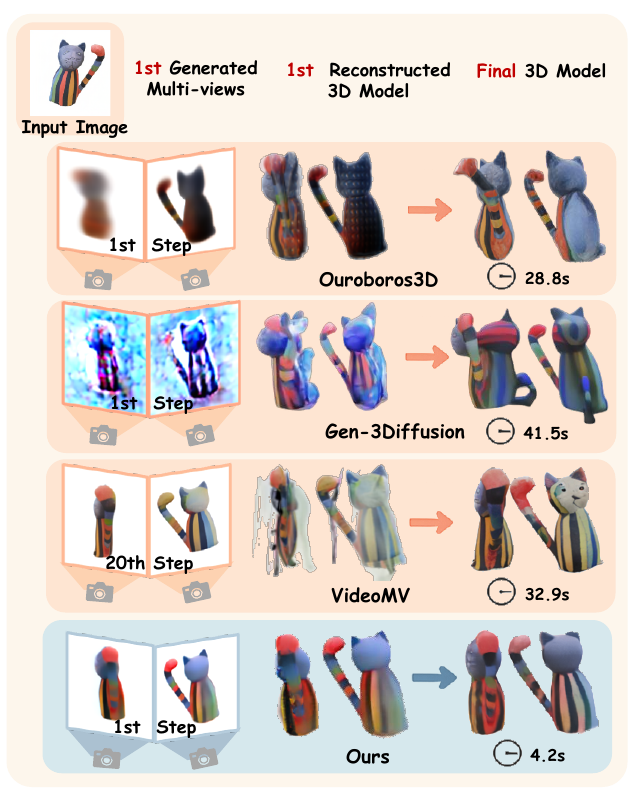}
    \vspace{-4mm}

     \caption{\textbf{The impact of CVC in 3D-aware sampling}. For Ouroboros3D \cite{wen2024ouroboros3d}, VideoMV \cite{zuo2024videomv}, and Gen-3Diffusion \cite{xue2024gen}, noise and lack of CVC affect 3D reconstructions. VideoMV applies feedback starting at the 20th step, so its results are shown after its first feedback. Our model, integrating Wonder3D~\cite{long2024wonder3d} and GeoLRM~\cite{zhang2024geolrm}, delivers better results with high time efficiency. %More comparisons are in the supplementary materials.
     }
    \vspace{-6mm}    

    \label{fig:noise_image}  
\end{figure}

\textbf{3D-Aware Sampling}. To address the above issues, 3D-aware sampling refines the 3D model by reimplementing the MVG model. As shown in Fig.~\ref{fig:flow} (b), during progressive denoising, the renderings of the reconstructed 3D outputs $\Tilde{\boldsymbol{X}}^{\pi}_{k+1}$ are re-noised at $t_{k}$ and fed back into the MVG model as the current denoising objective. This approach leverages explicit CVC from the reconstruction results to guide the sampling process, further enhancing the CVC of the MVG and producing a higher-quality 3D model.
Unlike existing 3D-aware sampling methods~\cite{wen2024ouroboros3d,zuo2024videomv,xue2024gen,xue2024human,tang2024cycle3d}, which generate noisy and inconsistent intermediates during denoising (Fig.~\ref{fig:noise_image}), further compounding inconsistencies in 3D generation, we propose a robust paradigm with a soft-hard alignment strategy to achieve more reliable and consistent 3D generation with enhanced CVC.

This strategy ensures consistent intermediates and progressive refinement, as shown in Fig.\ref{fig:teaser} and Fig.\ref{fig:noise_image}. Our method aligns both generation and reconstruction distributions with the ground truth (GT) distribution. Specifically, it ensures reconstructed multi-views align with a high-quality and manageable MVG distribution, while generated multi-views from the MVG align with the reconstruction model's high-CVC and generalizable domain. By alternating between these closely related near distributions, a stable 3D-aware sampling loop is established, allowing both models to excel within their respective domains.

 \begin{table*}[t]
    \centering
        \vspace{-5mm}

\fontsize{9}{6}\selectfont
    \begin{tabular}{@{}lccccccc@{}}
        \toprule
        Method & PSNR $\uparrow$ & SSIM $\uparrow$ & LPIPS $\downarrow$ & CLIP Sim. $\uparrow$ & FID $\downarrow$ & CVC $\uparrow$ & Time (s) $\downarrow$ \\
        \midrule
        LGM\textsuperscript{\textdagger}~\cite{tang2025lgm} & 18.010 & 0.893 & 0.145 & 0.859 & 113.945 & 4.316 & \textbf{\textcolor{red}{2.96}} \\
        Trellis~\cite{xiang2024structured} & 15.106 & 0.850 & 0.209 & 0.834 & 115.413 & - & 12.69 \\
        SyncDreamer~\cite{liu2023syncdreamer} & 18.423 & 0.885 & 0.153 & 0.787 & 270.902 & \textbf{\textcolor{blue}{5.668}} & 81.12 \\
        VideoMV~\cite{zuo2024videomv} & 18.042 & 0.802 & 0.151 & 0.825 & 127.095 & 5.363 & 32.93 \\
        Ouroboros3D~\cite{wen2024ouroboros3d} & \textbf{\textcolor{darkgreen}{21.069}} & \textbf{\textcolor{darkgreen}{0.902}} & \textbf{\textcolor{darkgreen}{0.117}} & \textbf{\textcolor{blue}{0.887}} & \textbf{\textcolor{blue}{102.379}} & 5.503 & 28.82 \\
        Gen-3Diffusion~\cite{xue2024gen} & \textbf{\textcolor{blue}{21.561}} & \textbf{\textcolor{blue}{0.904}} & \textbf{\textcolor{blue}{0.115}} & \textbf{\textcolor{darkgreen}{0.880}} & \textbf{\textcolor{darkgreen}{110.796}} & 5.023 & 41.52 \\
        \midrule
        \makecell[l]{Wonder3D~\cite{long2024wonder3d}+LGM} & 16.893 & 0.717 & 0.220 & 0.849 & 189.967 & 4.702 & 12.12 \\  
        \makecell[l]{AlignCVC (Ours)} & \textbf{\textcolor{red}{21.977}} & \textbf{\textcolor{red}{0.912}} & \textbf{\textcolor{red}{0.104}} & \textbf{\textcolor{red}{0.899}} & \textbf{\textcolor{red}{101.506}} & \textbf{\textcolor{red}{5.808}} & \textbf{\textcolor{darkgreen}{8.87}} \\  
        \cmidrule(r){1-1}
        \makecell[l]{Wonder3D+GeoLRM~\cite{zhang2024geolrm}} & 18.421 & 0.828 & 0.142 & 0.852 & 150.949 & 4.714 & 12.21 \\  
        \makecell[l]{AlignCVC (Ours)} & 19.693 & 0.860 & 0.128 & 0.881 & 128.379 & \textbf{\textcolor{darkgreen}{5.619}} & \textbf{\textcolor{blue}{4.17}} \\ 
        \cmidrule(r){1-1}
        \makecell[l]{Wonder3D+LaRa~\cite{chen2025lara}} & 16.490 & 0.719 & 0.250 & 0.823 & 220.770 & 4.671 & 5.01 \\  
        \makecell[l]{AlignCVC (Ours)} & 19.210 & 0.832 & 0.159 & 0.844 & 158.699 & 5.010 & 4.76 \\    
        \cmidrule(r){1-1}
        \makecell[l]{SV3D~\cite{voleti2024sv3d}+LGM} & 18.963 & 0.837 & 0.148 & 0.867 & 162.327 & 4.974 & 58.59  \\  
        \makecell[l]{AlignCVC (Ours)} & 20.483 & 0.892 & 0.120 & 0.876 & 117.220 & 5.522 & 17.48 \\  
        \cmidrule(r){1-1}
        \makecell[l]{SV3D+GeoLRM} & 17.872 & 0.816 & 0.142 & 0.836 & 150.977 & 4.873 & 55.17 \\  
        \makecell[l]{AlignCVC (Ours)} & 19.322 & 0.874 & 0.131 & 0.853 & 125.381 & 5.094 & 17.04 \\  
        \cmidrule(r){1-1}
        \makecell[l]{SV3D+LaRa} & 16.232 & 0.701 & 0.277 & 0.772 & 218.200 & 4.989 & 49.68 \\  
        \makecell[l]{AlignCVC (Ours)} & 18.270 & 0.847 & 0.142 & 0.841 & 127.942 & 4.995 & 16.87 \\   
        \bottomrule
    \end{tabular}
    \vspace{-2mm}

    \caption{\textbf{Comparison on single-image-to-3D generation.} 
    The top three results for each metric are highlighted in \textbf{\textcolor{red}{red}}, \textbf{\textcolor{blue}{blue}}, and \textbf{\textcolor{darkgreen}{green}}, respectively. LGM\textsuperscript{\textdagger} refers to the two-stage generation approach proposed in the original paper, which employs the pre-trained ImageDream~\cite{wang2023imagedream} model as the MVG, followed by the LGM reconstruction model.}
    \label{tab:duibi}
    \vspace{-5mm}
\end{table*}

\subsection{Soft-Aligned MVG}
\label{subsec:generation}

As mentioned above, in each step $k$, existing 3D-aware sampling methods often produce noisy multi-view predictions $\hat{\boldsymbol{X}}^\pi_k$, which are fed into the reconstruction model (see Fig.~\ref{fig:noise_image}). The reconstruction model then renders $\Tilde{\boldsymbol{X}}^\pi_k$ as feedback for the next denoising step. This recursive feedback mechanism perpetuates noise across iterations, limiting the MVG model’s ability to adapt to the directional shifts of the diffusion process and leading to a gradual degradation in the quality of 3D reconstruction. Consistent and robust guidance on the denoising direction is essential for enhancing the CVC of generated multi-view images. To address this, we propose shifting the MVG’s generation target to align $\{\hat{\boldsymbol{X}}^\pi_k\}^K_{k=1}$ with a broader GT-consistent distribution, rather than being restricted to fixed GT images in the dataset. This adjustment enables the MVG model to produce clearer images in the early stages and more effectively handle reconstruction results $\{\Tilde{\boldsymbol{X}}^\pi_k\}^K_{k=1}$, while maintaining flexibility and consistency with the GT distribution.

However, directly enforcing alignment with the GT distribution imposes overly strict constraints on individual views because the MVG model inherently learns CVC implicitly within its neural representations. Furthermore, using hard alignment methods, such as adversarial generative training, makes it easier for the discriminator to identify inconsistencies between the generated views and the GT distribution with CVC. This often results in mode collapse during training, as demonstrated in our experiments, further degrading the quality of generated multi-views.

Inspired by advances in Score Distillation~\cite{poole2022dreamfusion, wang2024prolificdreamer, yu2023text, ma2025scaledreamer}, we distill our MVG model using a pre-trained teacher that approximates the GT distribution. By leveraging the teacher's proximity to the GT and their shared distribution overlap, our model achieves \textbf{soft alignment} with the GT, which alleviates mode collapse caused by hard alignment and naturally aligns with diffusion training dynamics. Specifically, Score Distillation is employed to align the $\{\hat{\boldsymbol{X}}^\pi_k\}^K_{k=1}$ generated by the trainable MVG student with those predicted by the teacher. This alignment reduces noise in intermediate denoising stages, improving the stability of the generation process. Furthermore, aligning with the GT distribution, rather than strictly to the target GT images, enhances the model's generalization ability, particularly in adapting sampling guidance during 3D-aware sampling.

 Among recent distillation methods \cite{ma2025scaledreamer,wang2024prolificdreamer,poole2022dreamfusion,yu2023text,wang2023score,liang2023luciddreamer,wang2024esd}, we adopt Asynchronous Score Distillation (ASD) \cite{ma2025scaledreamer} as the distribution alignment objective due to its ability to produce stable gradients when training deep generative models. In this framework, the MVG teacher predicts the noise residual in $\hat{\boldsymbol{X}}_{k,t}^\pi$, which represents the $\hat{\boldsymbol{X}}_{k}^\pi$ generated by the MVG student diffused with Gaussian noise $\epsilon$ at timestep $t \in \{1, \dots, 1000\}$~\cite{ho2020denoising}. Let $\theta$ and $\phi$ denote the parameters of the MVG student and the MVG teacher, respectively. In ASD, given the input image $\boldsymbol{x}_c$ and camera parameters $\pi$ as conditions, and the $k$-step multi-view predictions $\hat{\boldsymbol{X}}^{\pi}_{k}$, the training gradient with respect to $\theta$ can be formulated as follows:
\begin{equation}
\setlength{\jot}{0pt} % 仅对这个公式的行间距生效
\begin{aligned}
\nabla_\theta \mathcal{L}_\text{Distill}(\hat{\boldsymbol{X}^{\pi}_{k}}) \triangleq 
\mathbb{E}_{t, \Delta t, \epsilon} \Big[ \omega(t) 
\Big( \epsilon_\phi(\hat{X}_{k,t}; t, x_c, \pi) \\
- \epsilon_\phi(\hat{X}_{k,t+\Delta t}; t+\Delta t, x_c, \pi) \Big)
\frac{\partial \hat{X}_k}{\partial \theta} \Big],
\end{aligned}
\label{eq:asd_loss}
\end{equation}
where $\omega(t)$ is a timestep-dependent weighting factor, and $\Delta t$ is a timestep shift proposed in ASD.
With the soft-aligned MVG, we achieve high-quality denoised results even in early diffusion steps, producing clearer multi-views with higher CVC. To this objective, we use pre-trained MVG teachers~\cite{long2024wonder3d,voleti2024sv3d} trained with multi-view renderings from the 3D object dataset~\cite{deitke2023objaverse} to ensure consistency between the distributions of generated and GT images. Instead of training from scratch, the student is initialized with the pre-trained teacher and fine-tuned by adding LoRA~\cite{hu2021lora} layers.

With the soft-alignment strategy, MVG student achieves clear multi-view images even at the first denoising step, ensuring outputs align closely with the GT and enhancing the generalizability of sampling guidance. Inspired by SD-Turbo \cite{sauer2024adversarial}, we limit the inference steps to as few as $K=4$, striking a trade-off between improving generation speed, avoiding excessive latent averaging \cite{huang2024mv}, and enabling effective 3D-aware sampling.

\subsection{Hard-Aligned Reconstruction}
\label{subsec:reconstruction}

Supervising the reconstruction model directly with GT multi-view images fails to resolve the distributional gap caused by inconsistent MVG-generated inputs, which often lack CVC. Our experiment also shows that such supervision severely degrades the model’s reconstruction ability, as the inconsistent inputs hinder effective multi-view aggregation. Forcing the model to strictly match GT views under these conditions leads to  blurry 3D reconstructions.

To address this, a distribution-based training objective for plausible reconstruction is necessary to ensure CVC across stages. Aligning with the GT distribution ensures the reconstructed model achieves consistency in both geometry and texture. While Score Distillation could theoretically align the distribution of $\{\Tilde{\boldsymbol{X}}^\pi_k\}^K_{k=1}$, the rendering process inherently enforces CVC, making a \textbf{hard alignment} strategy more effective. Specifically, adversarial generative learning aligns the reconstruction outputs directly with the GT distribution derived from renderings. This explicit supervision guarantees high-quality 3D models with robust CVC.

\begin{figure*}[h]
    \centering
    \vspace{-6mm}    

    \includegraphics[width=0.85\linewidth]{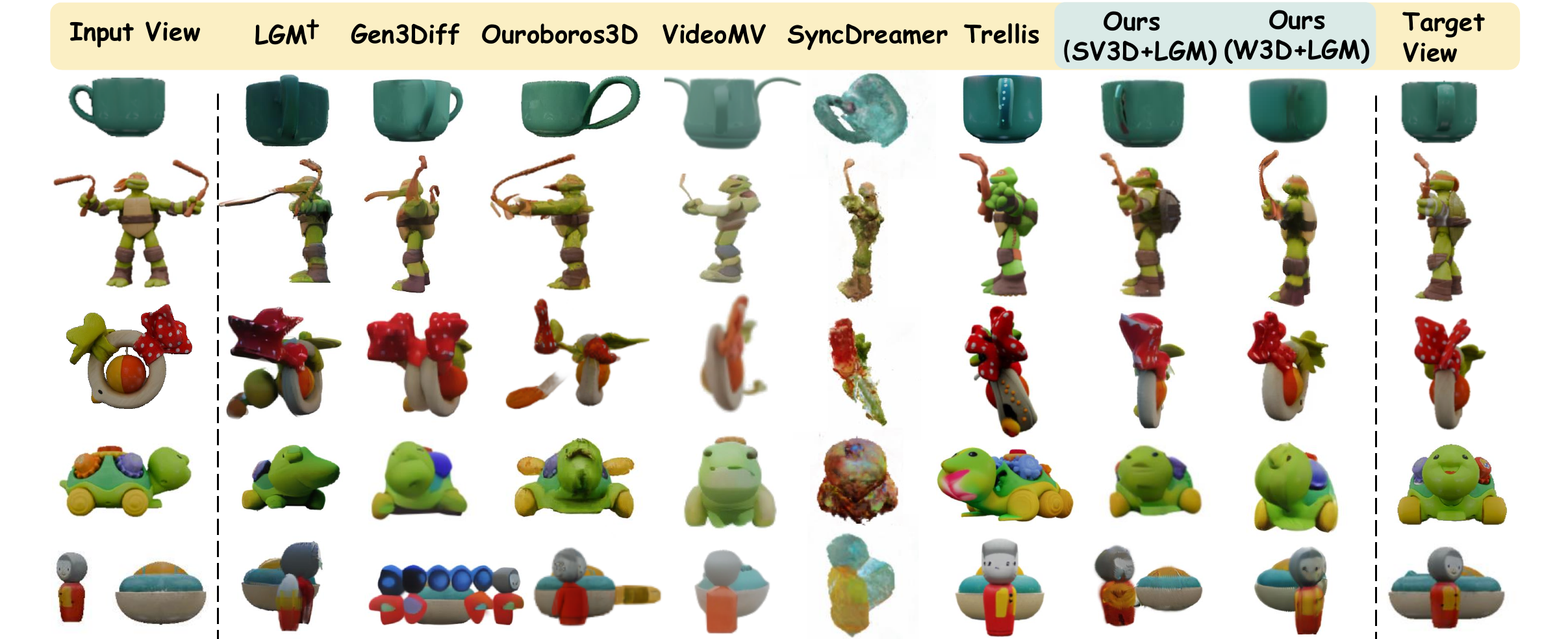}
    \vspace{-3mm}

    \caption{Comparison results on image-to-3D generation. Gen3Diff is short for Gen-3Diffusion.}     
    \label{fig:duibi} 
        \vspace{-3mm}    

\end{figure*}

We use the following adversarial objective to align the reconstruction results $\Tilde{\boldsymbol{X}}^{\pi}_{k}$ with the target distribution, providing a higher upper bound for alignment and enhancing robustness beyond strict regression losses:
\begin{equation}
\begin{aligned}
\mathcal{L}_{\text{GAN}} = \mathbb{E}_{X_{gt}}[\log D(X_{gt}^\pi)] + \mathbb{E}_{\Tilde{X}_{k}}[\log(1-D(\Tilde{X}_{k}^\pi))],
\end{aligned}
\label{eq:gan}
\vspace{-2mm}
\end{equation}
where $X_{gt}^\pi$ is the GT multi-views. We assist it with  a reconstruction loss for better performance:
\vspace{-3mm}

\begin{equation}
\begin{aligned}
\mathcal{L}_{\text{Recon}} = \| X_{gt}^\pi - {\Tilde{X}_{k}^\pi } \|^2_2.
\end{aligned}
\label{eq:recon}
\vspace{-2mm}
\end{equation}

In addition, since the reconstruction process uses multi-view images generated by the MVG as input, preserving the gradient connection between the two models facilitates more coordinated training. As a result, the objectives not only guide the MVG student to generate $\{\hat{\boldsymbol{X}}^\pi_k\}^K_{k=1}$, improving the alignment of $\{\Tilde{\boldsymbol{X}}^\pi_k\}^K_{k=1}$ with the GT distribution, but also directly supervise the final 3D model for quality. This process further enhances the reconstruction model through Score Distillation from the MVG. %Further details can be found in the supplementary materials.

\section{Experiments}
\subsection{Experimental Setup}

\textbf{Training Settings.}
We train our AlignCVC framework on the Gobjaverse dataset~\cite{qiu2024richdreamer}, which comprises multi-view renderings of 280K 3D objects sourced from Objaverse~\cite{deitke2023objaverse}. The resolution of the final results is set to 512×512 pixels for both training and testing. Our proposed method is trained for 300,000 iterations on an H20 GPU, utilizing the Adan~\cite{xie2024adan} optimizer with a learning rate of $2 \times 10^{-5}$. All computations are performed in 32-bit precision and implemented using the ThreeStudio~\cite{threestudio2023} framework.

\textbf{Testing Settings.} Following prior works, we evaluate our method with GSO~\cite{downs2022google} dataset. Tests are conducted on four orthogonal azimuths, and the evaluation metrics include PSNR, SSIM~\cite{wang2004image}, LPIPS~\cite{zhang2018unreasonable}, CLIP Similarity~\cite{radford2021learning}, and FID~\cite{kynkaanniemi2022role}.  
While these metrics primarily assess reconstruction quality against a fixed set of GT images, they inherently bias against models capable of generating novel yet plausible 3D-consistent content deviating from predefined GT. To address this limitation, we further evaluate the CVC of MVG outputs using the model introduced by DUSt3R~\cite{wang2024dust3r}, which predicts corresponding point maps between adjacent images. The CVC score is computed as the average confidence of the top 100 matched points based on their confidence scores. %More details of the CVC score can be found in the supplementary materials.

\textbf{Baseline MVG and Reconstruction Models.}
We prioritize MVG models that ensure camera control and generate high-fidelity multi-views for downstream tasks. Wonder3D~\cite{long2024wonder3d} and SV3D~\cite{voleti2024sv3d} are selected as MVG teachers for their ability to adjust camera parameters. Both are fine-tuned on the Gobjaverse dataset to better control elevation and restrict generated frames to orthogonal views.
For reconstruction, we adopt LGM~\cite{tang2025lgm} for its end-to-end prediction of 3DGS parameters from multi-views, along with GeoLRM~\cite{zhang2024geolrm} and LaRa~\cite{chen2025lara}, which project 2D features into 3D volumes to construct 3D models.
\begin{table*}[t]
 \centering
     %\vspace{-4mm}    

 \fontsize{9}{5}\selectfont
    \begin{tabular}{@{}ccccccc@{}}
    \toprule
Settings & \makecell{Loss\\ $\mathcal{L}_\text{Distill}$ $\mathcal{L}_\text{GAN}$ $\mathcal{L}_\text{Recon}$} & PSNR $\uparrow$ & SSIM $\uparrow$ & LPIPS $\downarrow$ & CVC $\uparrow$ & FID $\downarrow$ \\
\midrule

\makecell{Two-stage (Pre-trained Models)} & \makecell{\hspace{0.1cm}\textcolor{red}{\ding{55}}\hspace{0.3cm} \textcolor{red}{\ding{55}}\hspace{0.3cm} \textcolor{red}{\ding{55}}\hspace{0.1cm}} & 16.89 & 0.72 & 0.22 & 4.70 & 189.97 \\
\makecell{Two-stage (Train Recon.)} & \makecell{\hspace{0.1cm}\textcolor{red}{\ding{55}}\hspace{0.3cm} \textcolor{red}{\ding{55}}\hspace{0.3cm} \textcolor{green}{\ding{51}}\hspace{0.1cm}} & 18.24 \textcolor{red}{(+1.35)} & 0.66 & 0.16 \textcolor{red}{(-0.06)} & 4.70 & 194.44 \\
      \midrule
      \multicolumn{7}{c}{\textbf{+} 3D-Aware Sampling} \\
      \midrule
      \makecell{Pre-trained Models} & \makecell{\hspace{0.1cm}\textcolor{red}{\ding{55}}\hspace{0.3cm} \textcolor{red}{\ding{55}}\hspace{0.3cm} \textcolor{red}{\ding{55}}\hspace{0.1cm}} & 15.92 & 0.78 \textcolor{red}{(+0.06)} & 0.19 \textcolor{red}{(-0.03)} & 4.49 & 249.05 \\
      Fixed MVG & \makecell{\hspace{0.1cm}\textcolor{red}{\ding{55}}\hspace{0.3cm} \textcolor{green}{\ding{51}}\hspace{0.3cm} \textcolor{green}{\ding{51}}\hspace{0.1cm}} & 17.40 \textcolor{red}{(+0.51)} & 0.78 \textcolor{red}{(+0.06)} & 0.19 \textcolor{red}{(-0.03)} & 4.71 \textcolor{red}{(+0.01)} & 175.12 \textcolor{red}{(-14.85)} \\
      Fixed Recon. & \makecell{\hspace{0.1cm}\textcolor{green}{\ding{51}}\hspace{0.3cm} \textcolor{green}{\ding{51}}\hspace{0.3cm} \textcolor{green}{\ding{51}}\hspace{0.1cm}} & 19.05 \textcolor{red}{(+2.16)} & 0.86 \textcolor{red}{(+0.14)} & 0.12 \textcolor{red}{(-0.10)} & 5.12 \textcolor{red}{(+0.42)} & 142.16 \textcolor{red}{(-47.81)} \\
      
      \midrule
      \multicolumn{7}{c}{\textbf{+} Joint Training of Both MVG \& Recon.} \\
      \midrule
      \textbf{Ours} & \makecell{\hspace{0.1cm}\textcolor{green}{\ding{51}}\hspace{0.3cm} \textcolor{green}{\ding{51}}\hspace{0.3cm} \textcolor{green}{\ding{51}}\hspace{0.1cm}} & \textbf{21.98} \textcolor{red}{(+5.09)} & \textbf{0.91} \textcolor{red}{(+0.19)} & \textbf{0.10} \textcolor{red}{(-0.12)} & \textbf{5.81} \textcolor{red}{(+1.11)} & \textbf{101.51} \textcolor{red}{(-88.46)} \\
      w/o $\mathcal{L}_\text{Recon}$ & \makecell{\hspace{0.1cm}\textcolor{green}{\ding{51}}\hspace{0.3cm} \textcolor{green}{\ding{51}}\hspace{0.3cm} \textcolor{red}{\ding{55}}\hspace{0.1cm}} & 21.04 \textcolor{red}{(+4.15)} & 0.89 \textcolor{red}{(+0.17)} & 0.12 \textcolor{red}{(-0.10)} & 5.67 \textcolor{red}{(+0.97)} & 128.09 \textcolor{red}{(-61.88)} \\
      w/o MVG Align & \makecell{\hspace{0.1cm}\textcolor{red}{\ding{55}}\hspace{0.3cm} \textcolor{green}{\ding{51}}\hspace{0.3cm} \textcolor{green}{\ding{51}}\hspace{0.1cm}} & 18.28 \textcolor{red}{(+1.39)} & 0.84 \textcolor{red}{(+0.12)} & 0.14 \textcolor{red}{(-0.08)} & 5.02 \textcolor{red}{(+0.32)} & 150.52 \textcolor{red}{(-39.45)} \\
      w/o Recon. Align & \makecell{\hspace{0.1cm}\textcolor{green}{\ding{51}}\hspace{0.3cm} \textcolor{red}{\ding{55}}\hspace{0.3cm} \textcolor{green}{\ding{51}}\hspace{0.1cm}} & 20.84 \textcolor{red}{(+3.95)} & 0.88 \textcolor{red}{(+0.16)} & 0.12 \textcolor{red}{(-0.10)} & 5.31 \textcolor{red}{(+0.61)} & 124.38 \textcolor{red}{(-75.59)} \\
      w/o Distribution Align & \makecell{\hspace{0.1cm}\textcolor{red}{\ding{55}}\hspace{0.3cm} \textcolor{red}{\ding{55}}\hspace{0.3cm} \textcolor{green}{\ding{51}}\hspace{0.1cm}} & 18.16 \textcolor{red}{(+1.27)} & 0.83 \textcolor{red}{(+0.11)} & 0.16 \textcolor{red}{(-0.06)} & 4.51  & 187.05 \textcolor{red}{(-2.92)} \\

        \midrule
      \multicolumn{7}{c}{Impact of Alignment Strategy} \\
      \midrule
      
      Hard-aligned MVG & \makecell{\hspace{0.1cm}$\mathcal{L}_\text{GAN}^\text{MVG}$  \textcolor{green}{\ding{51}}\hspace{0.3cm} \textcolor{green}{\ding{51}}\hspace{0.1cm}} & 17.31 \textcolor{red}{(+0.42)} & 0.82 \textcolor{red}{(+0.10)} & 0.16 \textcolor{red}{(-0.06)} & 4.36  & 187.05 \textcolor{red}{(-2.92)} \\
     
      Soft-aligned Recon. & \makecell{\hspace{0.1cm}\textcolor{green}{\ding{51}}\hspace{0.3cm}$\mathcal{L}_\text{Distill}^\text{Recon.}$ \textcolor{green}{\ding{51}}\hspace{0.1cm}} & 16.69 &  0.80 \textcolor{red}{(+0.08)}& 0.18 \textcolor{red}{(-0.04)} & 3.97 & 191.17\\

       \midrule
      \multicolumn{7}{c}{Impact of Replacing Distillation Loss with Diffusion Loss} \\
      \midrule
  $\mathcal{L}_\text{Distill}^\text{MVG}$  → $\mathcal{L}_\text{Diff}^\text{MVG}$  & \makecell{\hspace{0.1cm}$\mathcal{L}_\text{Diff}^\text{MVG}$  \textcolor{green}{\ding{51}}\hspace{0.3cm} \textcolor{green}{\ding{51}}\hspace{0.1cm}} 
  & 16.39& 0.69  & 0.15 \textcolor{red}{(-0.07)} & 4.22 & 171.63 \textcolor{red}{(-18.34)} \\[0.3em]

+ w/o Recon. Align& \makecell{\hspace{0.1cm}$\mathcal{L}_\text{Diff}^\text{MVG}$  \textcolor{red}{\ding{55}}\hspace{0.3cm} \textcolor{green}{\ding{51}}\hspace{0.1cm}} &16.82 & 0.71 &0.15 \textcolor{red}{(-0.07)}  & 4.38 & 201.81 \\

      \bottomrule
    \end{tabular}

    \vspace{-2mm} % 减少表格下方的空间
 \caption{\textbf{Ablation studies.} Improvements over the pre-trained two-stage baseline (first row) are indicated in \textcolor{red}{red}. }
   \label{tab:ablation}
    \vspace{-6mm}    

\end{table*}

\textbf{Compared Methods}. We compare our method against recent open-source methods for single-image-to-3D generation. These include 3D-aware sampling approaches such as Ouroboros3D~\cite{wen2024ouroboros3d}, VideoMV~\cite{zuo2024videomv}, and Gen-3Diffusion~\cite{xue2024gen}. We also compare with two-stage methods, including LGM reconstruction model with its MVG model ImageDream~\cite{wang2023imagedream}, and SyncDreamer~\cite{liu2023syncdreamer}, which incorporates 3D features from intermediate renderings to produce more coherent multi-views, followed by NeRF-based reconstruction.  
Note that VideoMV and SyncDreamer, limited to 256×256 resolution, may benefit from smaller resolutions for higher PSNR and SSIM. We also compare with the 3D-based generator Trellis~\cite{xiang2024structured}.
%Due the limit of space, more experimental results can be found in the \textbf{supplementary materials}.%

\subsection{Results}
\textbf{Quantitative Comparison.} The quantitative results of competing methods are shown in Table~\ref{tab:duibi}. For two-stage image-to-3D generation, we combine pre-trained generation and reconstruction models, where multi-view images are first generated and then reconstructed into a 3D model. AlignCVC, as a plug-and-play paradigm, outperforms these combinations across all 3D metrics. Distribution alignment further enhances CVC performance, improving 3D model quality.
Table~\ref{tab:duibi} highlights the importance of pairing compatible generation and reconstruction models. For 3D generation, Wonder3D, optimized for orthogonal images, surpasses SV3D with better camera control. For reconstruction, LGM outperforms GeoLRM and LaRa by directly predicting 3DGS parameters, avoiding errors from projecting incomplete 2D images into 3D. By leveraging neural networks to aggregate multi-view features, LGM better adapts to distribution alignment, achieving significant gains.

\begin{figure*}[htp]
    \centering
    \includegraphics[width=1\linewidth]{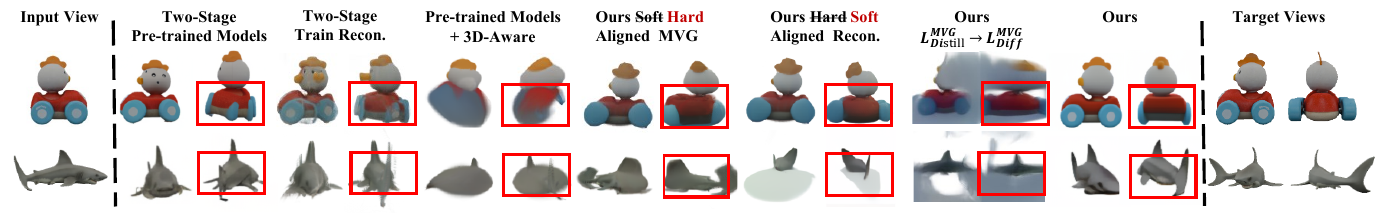}
    \vspace{-7mm}

    \caption{Ablation results of image-to-3D generation, evaluating the effect of different alignment strategies.}     
    \vspace{-6mm}
    \label{fig:ablation} 
    %\vspace{-8pt} % 减少表格下方的空间

\end{figure*}

Compared with existing methods in the upper panel of Table~\ref{tab:duibi}, AlignCVC integrating Wonder3D and LGM surpasses all SOTA methods. While Gen-3Diffusion demonstrates competitive generation quality, it requires significantly more time. On the other hand, the two-stage method LGM\textsuperscript{\textdagger} is the fastest but suffers from poor reconstruction metrics and CVC, prioritizing speed over quality. In contrast, AlignCVC achieves both high quality and fast inference speed, outperforming others in efficiency when using Wonder3D and LGM as baseline models.
In terms of the CVC metric, excluding Trellis, which generates 3D models without relying on MVG, LGM achieves the lowest scores due to the absence of 3D-aware sampling. SyncDreamer achieves the highest score by leveraging 3D features from multi-view images during sampling. Incorporating AlignCVC significantly enhances CVC performance, validating the effectiveness of our approach.

\textbf{Qualitative Comparison}. 
Fig.~\ref{fig:duibi} provides visual examples highlighting the superior CVC performance of our method compared to competing approaches, which exhibit inconsistencies and artifacts. The two-stage method, LGM\textsuperscript{\textdagger}, struggles with cross-view inconsistencies, leading to incorrect geometry and misaligned rotations. 3D-aware sampling methods like Gen-3Diffusion and Ouroboros3D suffer from noise propagation, producing redundant artifacts. Similarly, VideoMV and SyncDreamer, constrained by low resolution, display significant blurriness. Trellis, a 3D-native generator, preserves geometry well but has limited generalization to unseen data, resulting in deviations in color, semantics, and pose, ultimately lowering reconstruction metrics. Further results and discussions are in the supplementary materials.

%\subsection{Distribution Alignment}

\subsection{Ablation Study}

 %\vspace{-8pt}
In the ablation studies, we employ  Wonder3D as the MVG model and LGM as the reconstruction model. %Next we discuss the results in detail.

\textbf{The necessity of tuning both models.} 
We evaluate the \textbf{Two-stage} paradigm with pre-trained models as the baseline. We further train the reconstruction model to adapt to inconsistent multi-view images (serving as data augmentation) in the two-stage process, but this leads to blurrier reconstruction outputs. Additionally, directly applying 3D-aware sampling also fails to yield significant improvements. As shown in Fig.~\ref{fig:ablation}, these settings produce results inconsistent with the input, while 3D-aware sampling even degrades metrics like PSNR and FID.
This performance drop stems from the misalignment between the intermediate multi-views generated by the pre-trained MVG and the distributions of both the MVG teacher and ground truth (GT), emphasizing the need for distribution alignment. Experiments with \textbf{Fixed MVG} and \textbf{Fixed Recon.} further demonstrate that aligning either the reconstruction or MVG model significantly improves 3D-aware sampling performance.

\textbf{The necessity of distribution alignment.} 
With 3D-aware sampling and tuning of both MVG and reconstruction models, we investigate the impact of different distribution alignment objectives. With only the regression loss $\mathcal{L}_{\text{Recon}}$, as shown in \textbf{w/o Distribution Align} in Table~\ref{tab:ablation}, the improvement is limited. 
Building on this, applying alignment to either the MVG or the reconstruction model leads to better performance, as demonstrated in \textbf{w/o Recon. Align} and \textbf{w/o MVG Align}. However, when distribution alignment for the MVG model is removed, as shown in \textbf{w/o MVG Align}, the potential improvement is significantly hindered, indicating that MVG plays a more critical role in the sampling loop.
These results underscore the importance of effective MVG alignment objectives, which are essential for achieving optimal performance in 3D-aware sampling.

\textbf{The necessity of soft alignment for MVG model}. We use Score Distillation as the training objective for the MVG model, aligning its intermediate outputs with the distribution modeled by the MVG teacher. While this teacher distribution is derived from but not identical to the GT, directly aligning with the GT is another potential approach. However, combining hard alignment in the reconstruction model with adversarial training for the MVG model results in training collapse and performance degradation, as shown in \textbf{Hard-aligned MVG} in Table~\ref{tab:ablation} and Fig.~\ref{fig:ablation}. This is because the multi-view discriminator, relying on the CVC, easily distinguishes between generated and GT images, leading to unbalanced adversarial training and hindering the MVG model from effectively learning.
We also replace the distillation loss with diffusion loss~\cite{rombach2022high}, as shown in \ \textbf{Ours $\mathcal{L}_{\text{Distill}}^{\text{MVG}} \rightarrow \mathcal{L}_{\text{Diff}}^{\text{MVG}}$} in Table~\ref{tab:ablation} and Fig.~\ref{fig:ablation}, following the MVG training objective in Gen-3Diffusion. However, diffusion loss introduces significant noise to the rendered results, as observed in some Gen-3Diffusion cases, severely degrading the final 3D model quality. In this setting, reconstruction model alignment has minimal impact. %We further compare ASD with other Score Distillation methods in the supplementary materials.

 \textbf{The necessity of hard alignment for reconstruction model}. %We ablate the losses on the reconstruction model. 
 Using Score Distillation as the objective can lead to deteriorated performance, as shown in the rows of \textbf{Soft-aligned Recon.} in Table \ref{tab:ablation}, and the inconsistent output, such as the shark in Fig. \ref{fig:ablation}. It reveals that with explicit CVC, the reconstructed multi-views  are qualified for directly aligning with GT distributions. The ablation with \textbf{w/o $\mathcal{L}_\text{Recon}$} also shows that using the regression loss provides further improvement, because the reconstruction model is initially trained with this objective.

\vspace{-7pt}

\section{Conclusion}

We presented AlignCVC, a plug-and-play framework to enhance cross-view consistency (CVC) of single-image-to-3D generation by jointly post-training multi-view generation (MVG) and reconstruction models. AlignCVC addresses distribution misalignment, the key bottleneck in 3D-aware sampling, by aligning generated and reconstructed multi-view distributions with the GT distribution. Recognizing that CVC is implicit in MVG models but explicit in reconstruction models, we adopt a soft-hard alignment strategy: Score Distillation for MVG (soft) and adversarial training for reconstruction (hard). Experiments show AlignCVC improves performance across various MVG-reconstruction combinations, reducing 3D-aware sampling to 4 steps.

\textbf{Limitation} AlignCVC relies on two auxiliary networks (MVG teacher and discriminator) for distribution alignment, which increases GPU memory usage.

%\.\input{sec/X_suppl}

% your other appendix sections...
%\section{Acknowledgments}

\bigskip

\bibliography{main.bib}

\end{document}